\documentclass{article}

\PassOptionsToPackage{numbers,compress}{natbib}
\usepackage[preprint]{neurips_2026}
\usepackage[utf8]{inputenc}
\usepackage[T1]{fontenc}
\usepackage{hyperref}
\usepackage{url}
\usepackage{booktabs}
\usepackage{amsmath,amssymb,amsfonts,amsthm}
\usepackage{microtype}
\usepackage{xcolor}
\usepackage{graphicx}
\usepackage{multirow}
\usepackage{array}
\usepackage{algorithm}
\usepackage{algpseudocode}
\usepackage{placeins}

\graphicspath{{figures/}}

\newcommand{\E}{\mathbb{E}}
\newcommand{\R}{\mathbb{R}}

\newcommand{\gate}{g}
\newcommand{\Kset}{\mathcal K}

\title{Beyond ESG Scores: Learning Dynamic Constraints for Sequential Portfolio Optimization}

\author{%
  Xin Li \\
  Macquarie University \\
  \texttt{xin.li@mq.edu.au} \\
  \And
  Yan ke \\
  The University of Queensland \\
  \texttt{y.ke@uq.edu.au} \\
  \And
  Longbing Cao\thanks{Corresponding author.} \\
  Macquarie University \\
  \texttt{longbing.cao@mq.edu.au} \\
}

\begin{document}

\maketitle

\begin{abstract}
ESG-aware portfolio optimization is increasingly important for sustainable capital allocation, yet most learning-based methods still operationalize ESG by appending static scores to the policy observation or reward. This creates a mismatch for sequential control: ESG scores are noisy, provider-dependent, low-frequency, and temporally misaligned with sequential portfolio decisions, while financial evidence suggests that ESG is better treated as a portfolio preference, risk-exposure, or hedge dimension than as a robust alpha factor. We propose to impose ESG constraints without modifying the financial policy's observation or reward, using a Multimodal Action-Conditioned Constraint Field (MACF) that learns mechanism-specific ESG costs from point-in-time multimodal evidence and contemplated portfolio transitions. We then introduce MACF-X, a family of optimizer-specific adapters that converts MACF costs and uncertainties into native constrained-optimization interfaces through a shared slack- and uncertainty-aware pressure layer. Across multiple constraint-integration interfaces, MACF-X reduces tail ESG budget pressure while maintaining competitive financial performance.  Ablations show that this improvement depends on dynamic evidence inputs and three-head decomposition, while static ESG-score proxies are nearly indistinguishable from score-shuffled noise baselines.
\end{abstract}

\section{Introduction}

ESG investing seeks to align capital allocation with environmental responsibility, social impact, and corporate governance, and is an increasingly active area of AI for finance \citep{un2004whocares,cao2022aifinance,lim2024esgai,xu2024industrial}. Yet incorporating ESG constraints into a portfolio policy without turning sustainability into a noisy return signal remains challenging. Financial evidence suggests that ESG is better treated as a portfolio preference, risk exposure, or hedge dimension than as a robust stand-alone alpha factor \citep{pedersen2021responsible,pastor2021sustainable,jacobsen2019betterbeta,halbritter2015wages}. Widely used ESG scores are also noisy, provider-dependent, low-frequency, and sometimes revised retroactively \citep{berg2022aggregate,christensen2022virtue,berg2021rewriting,berg2022noise}. The core tension is that ESG-aware portfolio control requires point-in-time, action-relevant constraints, yet most existing pipelines still rely on static scores as the de facto ESG signal.

Traditional portfolio optimization already treats ESG as a constraint: screening, tilts, and explicit ESG constraints separate sustainability preferences from the financial objective \citep{branch2019guide,henriksson2019integrating,verheyden2016esg,qi2020imposing,chen2021social}, but these methods operate on static scores. Most learning-based ESG methods instead append ESG scores to the policy observation, reward, or multi-objective utility \citep{maree2022balancing,acero2024deep,garrido2025multiobjective}. A policy trained directly on such scores inherits measurement error, provider-specific bias, and temporal staleness---noise hard to disentangle from genuine sustainability signals. In sequential control, the key question is whether a contemplated portfolio move increases exposure to a currently salient ESG risk.

Standard safe RL methods assume that constraint costs are given by the environment \citep{achiam2017cpo,zhang2020focops,xu2021crpo}. In ESG-aware portfolio control, the relevant cost is latent: it must be inferred from sparse, asynchronous, and multimodal evidence, and it depends on the portfolio move under consideration. Buying into a firm under an active controversy, maintaining inherited exposure, and reducing exposure under peer spillover are different ESG-relevant decisions. The problem is therefore constrained optimization with learned, point-in-time, action-conditioned ESG costs.

We address this problem by learning ESG costs externally and injecting them as constraints, without modifying the policy's observation or reward. A \emph{Multimodal Action-Conditioned Constraint Field} (MACF) maps point-in-time multimodal evidence and contemplated portfolio transitions into mechanism-specific add-risk, hold-risk, and spillover-risk costs with uncertainty estimates. MACF therefore focuses on action-relevant ESG risk mechanisms rather than static asset scores.

Building on MACF, MACF-X exposes learned ESG costs and uncertainties through the native interfaces of constrained policy optimizers. Rather than modifying the policy observation or reward, MACF-X routes ESG information through a shared slack- and uncertainty-aware pressure layer. Depending on the backbone, this layer appears as adaptive cost pressure, a feasibility-switching signal, or a local safe-update geometry. Conceptually, MACF-X turns ESG integration from a feature-engineering choice inside the trading agent into an optimizer-side constraint interface.

\begin{figure}[t]
  \centering
  \includegraphics[width=\linewidth]{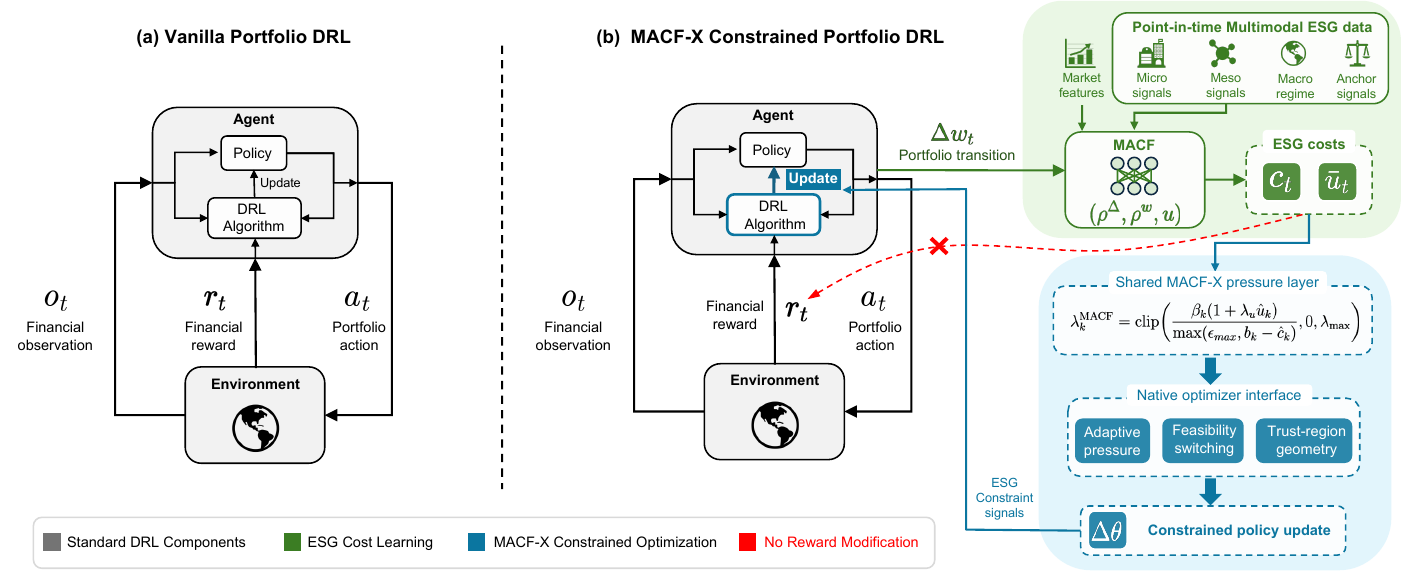}
  \caption{Comparison between vanilla DRL and MACF-X constrained portfolio DRL. (a) Vanilla DRL updates the policy using financial observations, portfolio actions, and financial rewards. (b) MACF-X preserves the financial-only policy observation and reward structure, while introducing ESG information through external MACF cost learning and constrained optimizer interfaces rather than reward modification.}
  \label{fig:architecture}
\end{figure}

Figure~\ref{fig:architecture} illustrates this separation: MACF-X preserves the financial policy loop while routing ESG evidence through external cost learning and constrained optimizer interfaces.

Our contributions are threefold:
\begin{enumerate}
  \item We formulate ESG-aware sequential portfolio optimization as a learned constrained-control problem, where ESG enters as point-in-time, action-conditioned costs rather than as static scores appended to the policy observation or reward.
  \item We introduce MACF, a multimodal action-conditioned constraint field that maps point-in-time evidence and contemplated portfolio transitions into decomposed add-risk, hold-risk, and spillover-risk costs with uncertainty estimates.
  \item We introduce MACF-X, a family of optimizer-specific adapters that inject MACF-derived ESG constraints into model-free portfolio optimization without modifying the policy observation or reward. MACF-X reduces tail ESG budget pressure while preserving financial performance; ablations show that gains require dynamic evidence and three-head decomposition rather than static ESG-score proxies.
\end{enumerate}

\section{Related work}

\paragraph{AI for ESG-aware portfolio optimization.}
ESG-aware portfolio optimization has traditionally used screening, tilts, and explicit objective- or constraint-based optimization, rather than learned point-in-time constraint representations \citep{branch2019guide,henriksson2019integrating,verheyden2016esg,qi2020imposing,chen2021social}. This literature suggests that ESG reshapes the opportunity set and efficient frontier more than it provides a stable stand-alone alpha: ESG enters naturally as a portfolio objective, preference, hedge, or constraint \citep{pedersen2021responsible,pastor2021sustainable,jacobsen2019betterbeta}, while empirical return evidence remains mixed and provider-dependent \citep{halbritter2015wages,ouchen2022less}. Recent RL work remains largely score-based, feeding ESG scores into the observation, reward, or multi-objective utility \citep{maree2022balancing,acero2024deep,garrido2025multiobjective}. Yet such scores suffer from cross-provider disagreement \citep{berg2022aggregate,christensen2022virtue}, retroactive revisions \citep{berg2021rewriting}, and noise in return tests \citep{berg2022noise}, leaving open the problem of constructing point-in-time, action-relevant ESG constraints for sequential portfolio control.

\paragraph{Constrained policy optimization.}
Classical CMDPs can be solved by dynamic programming or linear programming in small state--action spaces, but these methods do not scale to the continuous, high-dimensional policy classes used in modern portfolio control \citep{altman1999cmdp}. Since our goal is to attach learned ESG costs to existing portfolio DRL backbones, we focus on model-free constrained policy optimization, where methods differ mainly by how the cost signal enters the policy update: scalarized pressure, feasibility-driven switching or correction, and trust-region geometry.

\emph{Scalarization, penalty, and Lagrangian methods} reformulate the constrained problem by converting costs into scalar pressure on the reward-improvement update. Reward-constrained policy optimization and related primal--dual methods update the policy and Lagrange multipliers jointly, making them natural baselines for CMDPs with expected cost budgets \citep{tessler2019rcpo,stooke2020responsive}. Fixed-penalty and barrier variants avoid explicit dual learning by inserting fixed or structured cost terms into the optimization objective; examples include interior-point policy optimization with logarithmic barriers \citep{liu2020ipo} and P3O-style penalized policy improvement \citep{zhang2022p3o}. These methods are attractive because they preserve simple first-order updates, but their practical difficulty is that dual variables can oscillate, while fixed penalty coefficients become additional modeling choices and may optimize an objective that differs from the original constrained problem.

\emph{Feasibility-switching and first-order correction methods} use the cost signal to choose update direction rather than continuously scalarizing reward and cost. CRPO alternates between reward improvement and constraint reduction depending on policy feasibility \citep{xu2021crpo}; FOCOPS uses first-order policy-space updates to improve feasibility while retaining scalable optimization \citep{zhang2020focops}. This class is best understood as a protocol for switching or correcting policy updates under constraints.

\emph{Trust-region, projection, and safe-geometry methods} form a third interface by controlling local policy movement under constraints. TRPO supplies a local KL geometry for stable reward improvement \citep{schulman2015trpo}. CPO extends this trust-region template to CMDPs by adding cost constraints to the local policy update \citep{achiam2017cpo}, PCPO adds an explicit projection step back toward the feasible set \citep{yang2020pcpo}, and C-TRPO modifies the policy-space geometry so that trust regions better respect the safe policy set \citep{milosevic2025ctrpo}. These methods provide strong templates for constrained policy optimization, but still typically assume that the relevant cost signal is specified by the environment.

\section{Preliminaries}

\subsection{Sequential portfolio control with latent ESG evidence}

We consider daily portfolio control over $N$ risky assets and one cash asset. At time $t$, the full information available to the research system is
\begin{equation}
  x_t = \bigl(s_t^{\mathrm{fin}}, e_t^{\mathrm{esg}}\bigr),
\end{equation}
where $s_t^{\mathrm{fin}}$ denotes market, portfolio, and risk-control features, while $e_t^{\mathrm{esg}}$ denotes point-in-time ESG evidence such as firm-level incidents, peer pressure, macro-regime context, and anchor-confirmed events. The trading policy itself observes only
\begin{equation}
  o_t = s_t^{\mathrm{fin}},
\end{equation}
and samples a portfolio action
\begin{equation}
  a_t \sim \pi_\theta(\cdot \mid o_t).
\end{equation}
The action is converted into post-trade risky-asset weights $w_{t+1}\in\R^N$ and cash weight $w_{t+1}^{\mathrm{cash}}$, with $\Delta w_t=w_{t+1}-w_t$. The action converter enforces the same cash, single-name, and sector controls for all methods; Appendix~\ref{app:financial-protocol} gives the full portfolio-control protocol.

The financial training objective is intentionally ESG-free:
\begin{equation}
  J_r(\theta)
  =
  \E_{\pi_\theta}
  \left[
    \sum_{t=0}^{\infty}\gamma^t r_t
  \right],
\end{equation}
where
\begin{equation}
  r_t
  =
  \log(1+r_t^{\mathrm{port}})
  -
  c_{\mathrm{tx}}\sum_{i=1}^{N}|\Delta w_{i,t}|
  -
  \lambda_{\mathrm{dd}}\max(0, |dd_t|-d_0).
  \label{eq:financial-reward-new}
\end{equation}
The turnover and drawdown terms are financial frictions and risk-control penalties, shared across all experiments as detailed in Appendix~\ref{app:financial-protocol}. ESG does not enter the reward as alpha. Let $\Kset=\{\mathrm{add},\mathrm{hold},\mathrm{spill}\}$ denote the add-risk, hold-risk, and spillover-risk constraint mechanisms. ESG enters through learned average-cost constraints
\begin{equation}
  \bar J_{c_k}(\theta)
  =
  \E_{\pi_\theta}
  \left[
    \frac{1}{T}\sum_{t=0}^{T-1} c_t^{(k)}
  \right]
  \le b_k,
  \qquad k\in\Kset.
  \label{eq:cmdp-new}
\end{equation}
We use average costs because the ESG budgets are calibrated from empirical per-step cost distributions; this keeps the training estimates, budget thresholds, and reported ESG Violation metric on the same scale. The key challenge is that $c_t^{(k)}$ is not given by the environment: it must be learned from multimodal ESG evidence and aligned with portfolio actions.

\subsection{Optimizer interfaces considered}

Once learned ESG costs are available, model-free constrained portfolio optimizers mainly differ in where the cost signal enters the update. We use three interfaces: scalarized cost pressure for PPO-like methods, feasibility switching or correction for CRPO-style methods, and trust-region geometry for TRPO/CPO-style methods.

MACF-X instantiates these interfaces with the same MACF cost field, yielding adaptive pressure, a switching signal, or a local safe geometry depending on the backbone. Fixed-penalty PPO is retained only as a reward-shaping reference because it modifies the reward with a fixed coefficient rather than exposing ESG as an adaptive external constraint channel. Appendix~\ref{app:macfx-details} gives the optimizer-level update rules.

\section{Methodology}

\subsection{Framework overview}

The control loop in Figure~\ref{fig:architecture} starts from a financial DRL portfolio policy that observes market information and is updated by financial rewards. Our goal is to add ESG constraints without redefining either channel. MACF-X therefore routes ESG through a separate constraint path: MACF estimates action-conditioned ESG costs from point-in-time evidence for candidate portfolio transitions, and MACF-X maps the resulting portfolio-level costs and uncertainties into optimizer-specific ESG constraint signals for policy updates. This decomposition isolates two tasks that score-as-state or score-as-reward approaches conflate: learning an ESG cost signal not supplied by the environment, and using that signal without changing the policy observation space or financial objective.

\subsection{Learning action-conditioned ESG costs}

For asset $i$ at date $t$, define the structured ESG context
\begin{equation}
  H_{i,t}
  =
  \left(
  h_{i,t}^{\mathrm{micro}},
  h_{i,t}^{\mathrm{meso}},
  h_t^{\mathrm{macro}},
  h_{i,t}^{\mathrm{anchor}},
  h_{i,t}^{\mathrm{portfolio}},
  \Delta w_{i,t}
  \right),
  \label{eq:macf-context}
\end{equation}
where $h_{i,t}^{\mathrm{micro}}$ contains firm-level event memory, $h_{i,t}^{\mathrm{meso}}$ contains peer and industry pressure, $h_t^{\mathrm{macro}}$ contains market-regime context, $h_{i,t}^{\mathrm{anchor}}$ contains high-confidence evidence such as regulatory or official sources, and $h_{i,t}^{\mathrm{portfolio}}$ contains pre-trade portfolio context. The term $\Delta w_{i,t}=w_{i,t+1}-w_{i,t}$ is the signed target-weight change after action conversion, allowing MACF to distinguish increasing exposure to a firm under active ESG pressure from maintaining or reducing exposure without exposing ESG evidence to the financial policy.
Appendix~\ref{app:data-pipeline} describes the data construction pipeline used to assemble $H_{i,t}$.

MACF maps this context into head-wise output tensors:
\begin{equation}
  \left(
    \boldsymbol{\rho}_{i,t}^{\Delta},
    \boldsymbol{\rho}_{i,t}^{w},
    \mathbf u_{i,t}
  \right)
  =
  f_\phi(H_{i,t}),
  \qquad
  \boldsymbol{\rho}_{i,t}^{\Delta},
  \boldsymbol{\rho}_{i,t}^{w},
  \mathbf u_{i,t}\in[0,1]^{|\Kset|}.
  \label{eq:macf-output-new}
\end{equation}
The superscripts on $\boldsymbol{\rho}_{i,t}^{\Delta}$ and $\boldsymbol{\rho}_{i,t}^{w}$ denote exposure operators rather than MACF heads: $\rho_{i,t}^{\Delta,k}$ scores the cost of increasing exposure under mechanism $k$, $\rho_{i,t}^{w,k}$ scores the cost of maintaining post-trade exposure under mechanism $k$, and $u_{i,t}^{k}$ measures uncertainty for $k\in\Kset$.
MACF is intentionally structured: it uses cleaned multimodal evidence, categorical encodings, numerical risk-memory features, and a shared MLP with head-specific outputs. The model is trained before policy optimization using weak supervision from event-derived add-risk, hold-risk, spillover-risk, and uncertainty targets, and is frozen during DRL training; Appendix~\ref{app:macf-training} gives the full target construction, loss, and training protocol. The contribution is therefore not to scale the representation model, but to construct a point-in-time, action-conditioned ESG cost object that can be consumed directly by portfolio optimizers.

\begin{figure*}[!htbp]
  \centering
  \includegraphics[width=\textwidth]{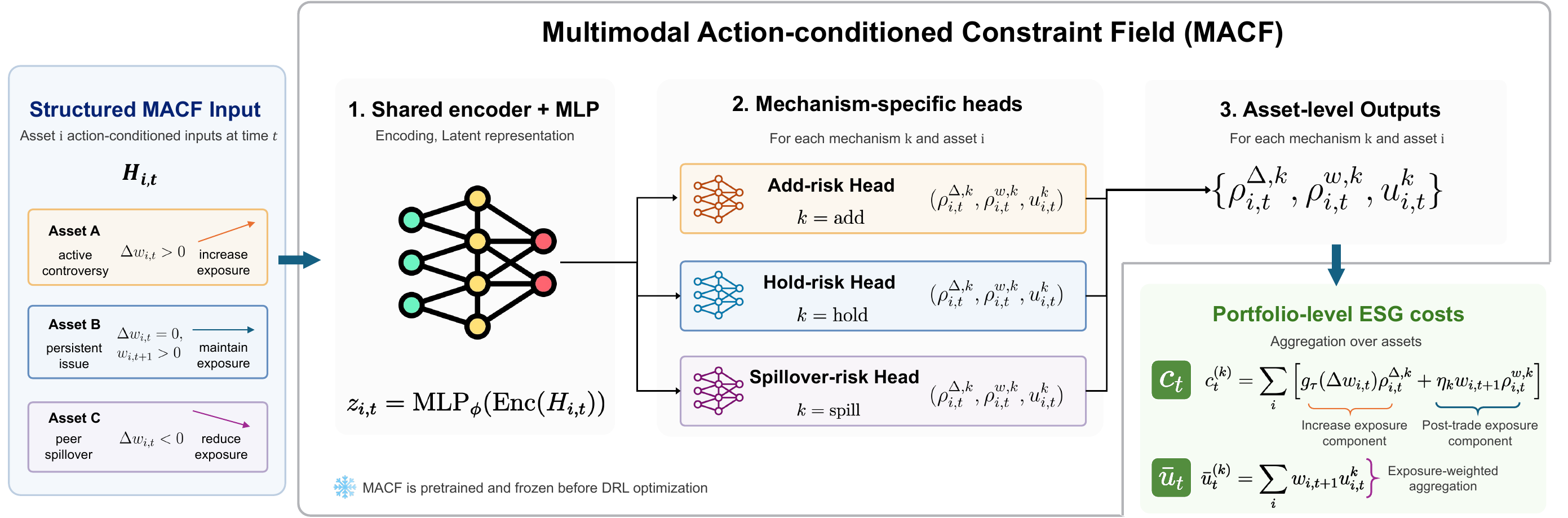}
  \caption{One-step construction of action-conditioned MACF costs. For each asset, the structured input $H_{i,t}$ combines point-in-time ESG evidence, pre-trade portfolio context, and the candidate weight change $\Delta w_{i,t}$. MACF maps this input through a shared encoder and mechanism-specific add-risk, hold-risk, and spillover-risk heads to produce asset-level components $(\rho_{i,t}^{\Delta,k},\rho_{i,t}^{w,k},u_{i,t}^{k})$. These components are aggregated across assets into portfolio-level ESG costs $\mathbf c_t$ and uncertainties $\bar{\mathbf u}_t$, which are passed to the MACF-X adapter.}
  \label{fig:macf-construction}
\end{figure*}

The key construction is asset-to-portfolio: MACF first scores each asset under add-risk, hold-risk, and spillover-risk mechanisms, then aggregates the resulting components using the candidate weight change and post-trade exposure. This yields portfolio-level ESG costs and uncertainties for MACF-X while preserving mechanism-specific asset-level semantics, as summarized in Figure~\ref{fig:macf-construction}.

\FloatBarrier

\subsection{Aggregating costs into portfolio constraints}

Given MACF outputs, we aggregate asset-level signals into portfolio costs:
\begin{equation}
  c_t^{(k)}
  =
  \sum_{i=1}^{N}
  \left[
    \gate_\tau(\Delta w_{i,t})
    \rho_{i,t}^{\Delta,k}
    +
    \eta_k w_{i,t+1}
    \rho_{i,t}^{w,k}
  \right],
  \qquad k\in\Kset.
  \label{eq:portfolio-cost-new}
\end{equation}
We use a nonnegative centered softplus gate
\begin{equation}
  \gate_\tau(x)
  =
  \left[
    \frac{1}{\tau}\log(1+\exp(\tau x))
    -
    \frac{\log 2}{\tau}
  \right]_+,
  \label{eq:centered-softplus}
\end{equation}
where $[z]_+=\max(z,0)$, $\tau>0$ controls smoothness, and $\eta_k\ge 0$ scales the exposure-maintenance component for mechanism $k$. The centering gives zero add-risk cost to an absence of trade, while the positive part prevents exposure reductions from contributing negative ESG cost. Portfolio-level uncertainty is
\begin{equation}
  \bar u_t^{(k)}
  =
  \sum_{i=1}^{N} w_{i,t+1}u_{i,t}^{k}.
  \label{eq:uncertainty-new}
\end{equation}
Budgets are calibrated per market from warm-start rollouts:
\begin{equation}
  b_k
  =
  Q_{q_k}\left(\mathcal C_k^{\mathrm{warm}}\right),
  \label{eq:budget-new}
\end{equation}
where $\mathcal C_k^{\mathrm{warm}}$ is the empirical distribution of per-step costs under a warm-start policy and $q_k$ is the head-specific calibration quantile. This makes the budget scale market-specific while keeping the experimental protocol fixed.

\subsection{Injecting constraints through MACF-X adapters}

MACF-X is an adapter family rather than a single optimizer. Its purpose is to translate MACF outputs into the native interface expected by different constrained-optimization methods. MACF produces the same abstract object for every optimizer: a head-wise, action-conditioned, point-in-time ESG cost field with an uncertainty estimate and a market-calibrated budget scale. This decouples constraint estimation from optimizer-specific update rules.

\paragraph{Lagrangian-style integration.}
For PPO-like methods, MACF-X supplies adaptive head-wise cost pressure through the shared slack-aware formula, yielding MACF-PPO without modifying the PPO policy architecture or financial objective.

\paragraph{Alternating-update integration.}
For CRPO-style first-order methods, MACF-X provides a mechanism-aware switching signal, yielding MACF-CRPO while preserving the first-order update profile of CRPO.

\paragraph{Trust-region integration.}
For TRPO/CPO-style methods, MACF-X reshapes the local trust-region geometry through learned ESG cost directions, producing MACF-TRPO and MACF-CPO.

The main text focuses on the shared pipeline: multimodal ESG evidence is converted into MACF costs, and MACF-X exposes those costs through the native interface of each optimizer. Appendix~\ref{app:macfx-details} gives the full update rules and implementation details for each instantiated optimizer.

\section{Experiments}

\subsection{Setup}

The main controlled study uses the US30 universe: thirty large, liquid U.S. equities balanced across six sectors; Appendix~\ref{app:universe-construction} explains the universe design and cross-market matching protocol. The split is 2021--2022 for training, 2023 for validation, and 2024 for test. MACF is trained on the training split, selected by validation loss, and then frozen before policy optimization. Adapter and optimizer hyperparameters are selected on the 2023 validation period and evaluated once on the 2024 test period. All methods use the same financial-only policy architecture, market-specific warm-start budget calibration, and portfolio-control protocol described in Appendix~\ref{app:financial-protocol}.

We group methods by optimizer interface: an unconstrained PPO reference, a fixed-penalty PPO reward-shaping baseline, and MACF-X adapters for scalarization, feasibility switching, and trust-region geometry. The code, processed data, and run configurations used in the experiments are provided at \url{https://anonymous.4open.science/r/SHUFHOSPNCUS}.

\subsection{Main US30 result}

\begin{table*}[!htbp]
  \centering
  \scriptsize
  \caption{US30 test-period comparison by optimizer interface. Return and turnover are percentages. ESG Violation is the largest realized ESG cost-to-budget ratio across heads and test dates; lower is better.}
  \label{tab:main-macfx}
  \begin{tabular}{lccccc}
    \toprule
    Method & Return & Sharpe & Calmar & Turnover & ESG Violation \\
    \midrule
    \multicolumn{6}{c}{\textit{PPO Reference and Fixed-Penalty Baseline}} \\
    \midrule
    PPO & 24.13\% & 2.34 & 4.54 & 14.13\% & 2.36 \\
    PPO+Penalty & 31.34\% & 2.71 & 6.39 & 15.09\% & \textbf{1.84} \\
    \midrule
    \multicolumn{6}{c}{\textit{Scalarization / Lagrangian Interface}} \\
    \midrule
    PPO-Lagrangian & 31.27\% & \textbf{2.79} & \textbf{6.59} & 13.69\% & \textbf{1.81} \\
    \textbf{MACF-PPO} & \textbf{31.60\%} & 2.68 & 6.00 & 13.56\% & 2.05 \\
    \midrule
    \multicolumn{6}{c}{\textit{First-Order Feasibility Switching Interface}} \\
    \midrule
    CRPO & 31.04\% & 2.59 & 5.72 & \textbf{11.71\%} & 1.80 \\
    \textbf{MACF-CRPO} & \textbf{31.17\%} & \textbf{2.72} & \textbf{6.35} & 14.87\% & \textbf{1.74} \\
    \midrule
    \multicolumn{6}{c}{\textit{Trust-Region / Safe Geometry Interface}} \\
    \midrule
    TRPO & 23.53\% & 2.31 & 4.52 & 8.97\% & 1.97 \\
    CPO & 23.89\% & 2.34 & 4.62 & \textbf{8.03\%} & 1.97 \\
    MACF-TRPO & 23.89\% & 2.39 & 4.29 & 13.57\% & \textbf{1.85} \\
    \textbf{MACF-CPO} & \textbf{24.48\%} & \textbf{2.40} & \textbf{4.91} & 9.06\% & 1.90 \\
    \bottomrule
  \end{tabular}%
\end{table*}

Table~\ref{tab:main-macfx} reports interface-specific trade-offs rather than a single dominant method. ESG Violation is $\max_{t,k} c_t^{(k)}/b_k$, so values above one indicate a budget exceedance and lower is better. Relative to unconstrained PPO, all constrained methods reduce tail ESG budget pressure. Within the scalarization/Lagrangian interface, MACF-PPO has the highest return but a higher ESG Violation value than PPO-Lagrangian. Within feasibility switching, MACF-CRPO improves return, Sharpe, Calmar, and ESG Violation relative to CRPO. Within trust-region methods, MACF-TRPO gives the lowest ESG Violation value, whereas MACF-CPO gives the strongest financial metrics among trust-region variants.

\subsection{Cross-market validation}

To test whether the learned-cost integration story is market-specific, we additionally evaluate the same protocol on the EU30 universe matched in Appendix~\ref{app:universe-construction}. The EU30 study keeps the reward, policy architecture, minimum-invested ratio, and budget-calibration protocol fixed, but recalibrates budgets on the EU30 training environment because ESG cost scales differ across universes and market structures. We interpret EU30 as a cross-market robustness check rather than as the primary controlled study. The relevant question is whether MACF-X preserves or improves the return--safety trade-off within each optimizer interface under the same protocol.

\begin{table*}[!htbp]
  \centering
  \scriptsize
  \caption{EU30 cross-market validation under the aligned reward, budget-calibration, and minimum-invested-ratio protocol. Return and turnover are percentages. ESG Violation is the largest realized ESG cost-to-budget ratio across heads and test dates; lower is better.}
  \label{tab:eu30-main}
  \begin{tabular}{lccccc}
    \toprule
    Method & Return & Sharpe & Calmar & Turnover & ESG Violation \\
    \midrule
    \multicolumn{6}{c}{\textit{PPO Reference and Fixed-Penalty Baseline}} \\
    \midrule
    PPO & 10.86\% & 1.04 & 1.26 & 19.67\% & 8.69 \\
    PPO+Penalty & 11.89\% & 1.17 & 1.40 & 18.97\% & \textbf{6.68} \\
    \midrule
    \multicolumn{6}{c}{\textit{Scalarization / Lagrangian Interface}} \\
    \midrule
    PPO-Lagrangian & 10.87\% & \textbf{1.10} & \textbf{1.28} & 16.63\% & \textbf{4.73} \\
    \textbf{MACF-PPO} & \textbf{10.90\%} & 1.07 & 1.27 & \textbf{16.60\%} & 5.00 \\
    \midrule
    \multicolumn{6}{c}{\textit{First-Order Feasibility Switching Interface}} \\
    \midrule
    CRPO & \textbf{10.86\%} & 1.04 & 1.26 & 19.67\% & 8.69 \\
    \textbf{MACF-CRPO} & 10.37\% & \textbf{1.07} & \textbf{1.27} & \textbf{16.54\%} & \textbf{4.55} \\
    \midrule
    \multicolumn{6}{c}{\textit{Trust-Region / Safe Geometry Interface}} \\
    \midrule
    TRPO & 10.76\% & 1.08 & 1.30 & \textbf{17.70\%} & 6.20 \\
    CPO & 10.76\% & 1.08 & 1.30 & \textbf{17.70\%} & 6.20 \\
    \textbf{MACF-TRPO} & \textbf{11.80\%} & \textbf{1.17} & \textbf{1.42} & 19.56\% & 6.67 \\
    MACF-CPO & 11.30\% & 1.15 & 1.42 & 20.83\% & \textbf{5.17} \\
    \bottomrule
  \end{tabular}%
\end{table*}

Table~\ref{tab:eu30-main} gives a cross-market robustness check. The scalarization/Lagrangian comparison is closest to US30: MACF-PPO slightly improves return over PPO-Lagrangian, with ESG Violation well below unconstrained PPO. In feasibility switching, MACF-CRPO lowers ESG Violation and turnover at the cost of a small return decrease. In the trust-region interface, CPO coincides with TRPO because the recovery branch is not activated; MACF-TRPO has the best financial metrics, while MACF-CPO has the lowest ESG Violation among trust-region variants. Thus, the same MACF-X protocol transfers to EU30 without architectural changes, although the magnitude and direction of the return--safety trade-off remain market dependent.

\subsection{MACF field-quality ablations}

The main policy tables evaluate whether learned ESG costs can be attached to constrained portfolio optimizers. We next isolate the representation question: whether MACF itself learns an action-relevant constraint field. This distinction is important because a weaker or flatter constraint field can sometimes release the portfolio optimizer and improve raw financial return, which is not evidence that the field is better. We therefore evaluate ablations at the field level, before policy optimization, to test whether dynamic evidence inputs and mechanism-specific heads capture ESG risk structure that a weaker or scalar field would miss.

For each MACF variant, we train on 2021--2022, select by 2023 validation loss, and evaluate the frozen field on the 2024 test split. We report RMSE against the weak-supervision targets, $\mathrm{AUC}_{90}$ for ranking top-decile target-risk samples, and top-10 lift, defined as the mean target risk among the top 10\% scored samples divided by the unconditional mean target risk. Metrics are computed separately for add-risk, hold-risk, and spillover-risk and then averaged.

\begin{table*}[!htbp]
  \centering
  \scriptsize
  \caption{US30 MACF input ablation on the 2024 test split. Lower RMSE and validation loss indicate better calibration; higher $\mathrm{AUC}_{90}$ and top-10 lift indicate better high-risk ranking.}
  \label{tab:macf-input-ablation}
  \setlength{\tabcolsep}{3.5pt}
  \begin{tabular}{lccccccc}
    \toprule
    MACF input variant & Mean RMSE & Mean $\mathrm{AUC}_{90}$ & Mean Lift & Add RMSE & Hold RMSE & Spill RMSE & Val. Loss \\
    \midrule
    Full MACF & \textbf{0.039} & \textbf{0.954} & 1.650 & \textbf{0.065} & 0.038 & \textbf{0.013} & \textbf{0.003} \\
    Remove portfolio context & 0.039 & 0.948 & 1.629 & 0.069 & \textbf{0.033} & 0.015 & \textbf{0.003} \\
    Remove micro event cards & 0.055 & 0.885 & 1.447 & 0.099 & 0.051 & 0.015 & 0.023 \\
    Remove firm memory & 0.077 & 0.925 & \textbf{1.658} & 0.075 & 0.136 & 0.020 & 0.022 \\
    Remove meso peer statistics & 0.070 & 0.868 & 1.555 & 0.064 & \textbf{0.033} & 0.113 & 0.010 \\
    Remove macro regimes & 0.056 & 0.940 & 1.606 & 0.073 & 0.040 & 0.056 & 0.005 \\
    Remove action context & 0.060 & 0.951 & 1.645 & 0.128 & 0.037 & 0.015 & 0.008 \\
    Remove all dynamic ESG evidence & 0.150 & 0.544 & 1.031 & 0.125 & 0.195 & 0.129 & 0.076 \\
    \bottomrule
  \end{tabular}%
\end{table*}

Table~\ref{tab:macf-input-ablation} gives a mechanism-aligned degradation pattern. Removing micro event cards primarily weakens add-risk identification; removing firm memory mainly affects hold-risk calibration; removing meso peer statistics sharply degrades spillover-risk prediction. Removing action context also harms add-risk calibration, consistent with add-risk being an exposure-increase mechanism rather than a ticker-level property. Without dynamic ESG evidence, mean RMSE rises from $0.039$ to $0.150$, $\mathrm{AUC}_{90}$ falls from $0.954$ to $0.544$, and top-10 lift falls from $1.65$ to nearly $1$. The learned field therefore depends on point-in-time ESG evidence rather than only on broad market or portfolio regularities.

\begin{table*}[!htbp]
  \centering
  \scriptsize
  \caption{US30 MACF head-decomposition ablation on the 2024 test split. Shared-scalar variants compress add-risk, hold-risk, and spillover-risk before evaluation against mechanism-specific targets.}
  \label{tab:macf-head-ablation}
  \setlength{\tabcolsep}{4pt}
  \begin{tabular}{lcccccc}
    \toprule
    Field structure & Mean RMSE & Mean $\mathrm{AUC}_{90}$ & Mean Lift & Add RMSE & Hold RMSE & Spill RMSE \\
    \midrule
    Full three-head MACF & \textbf{0.039} & \textbf{0.954} & \textbf{1.650} & \textbf{0.065} & \textbf{0.038} & \textbf{0.013} \\
    Shared predicted scalar & 0.233 & 0.803 & 1.495 & 0.270 & 0.308 & 0.120 \\
    Oracle shared scalar & 0.234 & 0.833 & 1.575 & 0.270 & 0.311 & 0.120 \\
    \bottomrule
  \end{tabular}%
\end{table*}

Table~\ref{tab:macf-head-ablation} evaluates whether the three MACF heads can be replaced by a single scalar risk. Compressing the learned field into a shared scalar increases mean RMSE by roughly a factor of six and reduces high-risk ranking quality. The oracle shared scalar, which averages the three target labels directly, also remains far worse than the three-head field. This indicates that the gap is not only due to prediction error; the three mechanisms encode different action semantics.

\subsection{Static ESG-score ablation}

If static ESG scores already provided meaningful point-in-time constraint signals, learning a dynamic evidence-based field would be unnecessary. This ablation tests that alternative by comparing three cost sources on US30 under the same policy architecture, financial protocol in Appendix~\ref{app:financial-protocol}, warm-start budget-calibration procedure, and five random seeds: (i) the full MACF field learned from multimodal point-in-time evidence, (ii) a static ESG-score proxy, and (iii) a shuffled static-score proxy that preserves the score distribution and point-in-time coverage but randomly permutes the ticker-to-score assignment. Because this diagnostic isolates the cost source under the same optimizer configurations, we evaluate it over five random seeds to verify that the near-zero Static--Shuffled gaps are not an artifact of a single training run.

\begin{table*}[!htbp]
  \centering
  \scriptsize
  \caption{US30 cost-source diagnostic over five seeds. Values are mean absolute gaps over the tested optimizer set, reported as mean $\pm$ standard deviation across seeds. Return and turnover gaps are percentage points; ESG Violation gaps are dimensionless.}
  \label{tab:static-score-ablation}
  \begin{tabular}{lccc}
    \toprule
    Metric & Mean $|\mathrm{MACF}-\mathrm{Static}|$ & Mean $|\mathrm{MACF}-\mathrm{Shuffled}|$ & Mean $|\mathrm{Static}-\mathrm{Shuffled}|$ \\
    \midrule
    Return (pp) & $1.57\pm0.62$ & $1.57\pm0.62$ & $0.00\pm0.00$ \\
    Sharpe & $0.09\pm0.05$ & $0.09\pm0.05$ & $0.00\pm0.00$ \\
    Calmar & $0.39\pm0.18$ & $0.39\pm0.18$ & $0.00\pm0.00$ \\
    Turnover (pp) & $1.21\pm0.25$ & $1.21\pm0.25$ & $0.00\pm0.00$ \\
    ESG Violation & $0.72\pm0.10$ & $0.56\pm0.19$ & $0.27\pm0.22$ \\
    \bottomrule
  \end{tabular}%
\end{table*}

Table~\ref{tab:static-score-ablation} is a diagnostic rather than a raw-return ranking. Static ESG scores and ticker-shuffled static scores produce identical financial gaps across seeds, while MACF differs more substantially from both controls in financial outcomes and tail budget pressure. This does not imply that MACF mechanically maximizes raw return relative to every static-score regularizer; it indicates that preserving only the marginal static-score distribution does not reproduce the constraint landscape learned from point-in-time evidence.

\section{Conclusions}

In this paper, we introduced MACF, a multimodal action-conditioned ESG constraint field, and MACF-X, a family of optimizer-specific adapters for injecting learned ESG constraints into portfolio DRL without modifying the financial reward or policy observation. MACF converts point-in-time ESG evidence and contemplated portfolio transitions into mechanism-specific add-risk, hold-risk, and spillover-risk costs with uncertainty estimates. MACF-X then exposes these costs through native constrained-optimization interfaces, including scalarization, feasibility switching, and trust-region interfaces. Across US30 and EU30, the results show that ESG constraints can be integrated with multiple model-free portfolio optimization backbones while reducing tail ESG budget pressure and preserving competitive financial performance.

More broadly, we argue that ESG-aware portfolio control should not treat sustainability as a noisy reward-side alpha proxy. A more promising direction is to study when ESG signals behave as dynamic sustainability-risk hedges across markets and asset classes, and how such signals can be incorporated as constraints rather than as return predictors. MACF-X represents one step in this direction: it shows that existing model-free portfolio optimizers can absorb learned ESG constraints through adapter-based interfaces while preserving the financial policy objective. Future work could extend this idea to cross-market and cross-asset hedging settings, test additional portfolio DRL and safe-RL backbones, and improve multimodal ESG evidence extraction with large models and temporal analytics. Progress in this area would also benefit from public, auditable, point-in-time ESG datasets and shared evaluation protocols. Together, these directions can help move ESG-aware AI from static score integration toward dynamic, constraint-based sustainable portfolio control.

\bibliographystyle{unsrtnat}
\bibliography{references}

\appendix

\section{Point-in-time ESG data construction pipeline}
\label{app:data-pipeline}

We construct the structured MACF input context $H_{i,t}$ through a point-in-time pipeline that separates financial state variables, firm-level ESG evidence, peer aggregation, macro regimes, and candidate weight changes. The processed research-use data used in the experiments are included in the anonymized repository; a more detailed and extensible data-construction pipeline will be released in future dataset-oriented work. This is important for the broader ESG-aware portfolio optimization community, where progress is currently limited by the scarcity of public, reliable, point-in-time multimodal ESG data.

\begin{figure*}[!htbp]
  \centering
  \includegraphics[width=\textwidth]{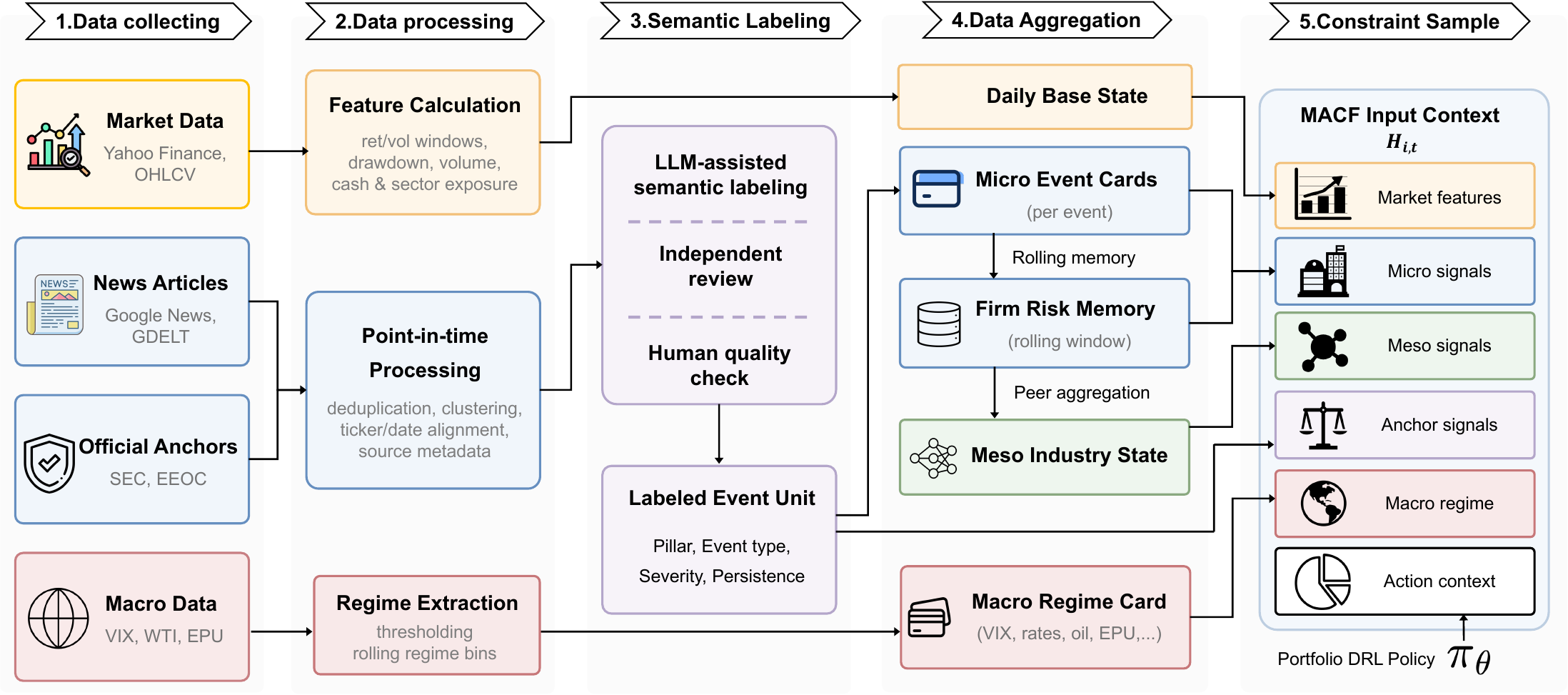}
  \caption{Point-in-time ESG data construction pipeline. Market data are converted into daily market and portfolio features; news and official anchors are deduplicated, aligned by ticker and date, semantically labeled, and aggregated into micro event cards, firm risk memory, meso peer states, and anchor signals; macro series are converted into regime cards. The resulting components form the structured MACF input context $H_{i,t}$, while the candidate weight change is supplied by the portfolio DRL policy.}
  \label{fig:data-pipeline}
\end{figure*}
\FloatBarrier

The pipeline separates financial state construction from ESG evidence construction. Market prices and portfolio records produce daily base features such as return and volatility windows, drawdown, volume, cash exposure, and sector exposure. These variables support the financial and portfolio-context components of $H_{i,t}$ but do not define ESG costs by themselves.

We build firm-level ESG evidence from news articles and official anchors such as regulatory or official-source material. We deduplicate and cluster raw records, align them to ticker-date pairs, and convert them into point-in-time event units with source metadata. A semantic labeling stage assigns event attributes such as ESG pillar, event type, severity, and persistence, with independent review and human quality checks used to reduce noisy or ambiguous labels. Labeled events are then aggregated into micro event cards and rolling firm risk memory.

The remaining components capture broader transmission channels. Firm-level event memory is aggregated across peers to construct meso industry states, while macro variables such as volatility, rates, oil, and policy uncertainty are discretized into macro regime cards. Officially anchored events also contribute anchor signals, allowing MACF to distinguish ordinary news flow from higher-confidence evidence. Together, these components provide MACF with point-in-time multimodal context while leaving the portfolio policy's financial observation space unchanged.

\section{MACF-X implementation details}
\label{app:macfx-details}

This appendix specifies how the common MACF constraint field is attached to the optimizer interfaces discussed in the main text. The goal is to make clear that MACF-X is not a collection of unrelated optimizers. Each variant consumes the same rollout object: reward advantages, head-wise MACF cost advantages, head-wise per-step cost estimates, head-wise uncertainties, and market-calibrated budgets. The variants differ only in the native interface through which this object enters the policy update.

\subsection{Shared rollout quantities and notation}

For each rollout collected under the current policy $\pi_{\theta_{\mathrm{old}}}$, the environment returns the financial reward $r_t$, the MACF cost vector
\begin{equation}
  \mathbf c_t
  =
  \left(c_t^{\mathrm{add}},c_t^{\mathrm{hold}},c_t^{\mathrm{spill}}\right),
\end{equation}
where the superscripts are compact notation for add-risk, hold-risk, and spillover-risk. The environment also returns the portfolio-level uncertainty vector $\bar{\mathbf u}_t$ defined in Eq.~\eqref{eq:uncertainty-new}. We estimate reward and cost advantages with the same trajectory data:
\begin{equation}
  A_t^r,
  \qquad
  A_t^{c,k},
  \qquad
  k\in\Kset.
\end{equation}
Following the average-cost constraint in Eq.~\eqref{eq:cmdp-new}, update decisions use rollout-level per-step estimates:
\begin{equation}
  \hat c_k
  =
  \frac{1}{T}\sum_{t=0}^{T-1} c_t^{(k)},
  \qquad
  \hat u_k
  =
  \frac{1}{T}\sum_{t=0}^{T-1} \bar u_t^{(k)}.
  \label{eq:appendix-vhat}
\end{equation}
This keeps the estimate and the budget on the same scale while preserving the CMDP interpretation of ESG as a constraint rather than as reward-side alpha.

The common MACF-X pressure weight is
\begin{equation}
  \lambda_k^{\mathrm{MACF}}
  =
  \mathrm{clip}
  \left(
    \frac{
      \beta_k(1+\lambda_u \hat u_k)
    }{
      \max(\epsilon_{\mathrm{safe}}, b_k-\hat c_k)
    },
    0,\lambda_{\max}
  \right),
  \label{eq:macfx-lambda}
\end{equation}
where $\beta_k$ is a head-specific scale, $\lambda_u$ controls uncertainty amplification, $\epsilon_{\mathrm{safe}}$ prevents division by zero, and $\lambda_{\max}$ prevents numerical explosion near the boundary. The same formula is used across MACF-PPO, MACF-CRPO, MACF-TRPO, and MACF-CPO. This is the main point of framework consistency: MACF always supplies the same head-wise cost field, and MACF-X always converts the field into slack- and uncertainty-aware pressure. The only difference is whether that pressure appears as a scalarized advantage, a switching rule, or a local geometry.

In the reported US30 configuration, MACF-PPO and MACF-CRPO use $\beta=(1,1,1)$, $\lambda_u=1$, $\epsilon_{\mathrm{safe}}=10^{-8}$, and $\lambda_{\max}=100$; MACF-CRPO uses $\lambda_{\mathrm{thr}}=50$. MACF-TRPO and MACF-CPO use $\beta=(0.25,1,1)$, $\lambda_u=0.5$, $\epsilon_{\mathrm{safe}}=10^{-8}$, and $\lambda_{\max}=30$, with a KL radius $\delta=0.03$ and recovery radius $\delta_{\mathrm{rec}}=0.005$. These constants are chosen on the validation period and kept fixed for test evaluation.

We instantiate MACF-X on three representative constrained-optimization interfaces. PPO-style Lagrangian updates represent scalarized pressure and primal--dual methods, CRPO represents first-order feasibility switching, and TRPO/CPO represent trust-region safe geometry. These interfaces are standard in constrained RL and expose different entry points for the same learned constraint object: scalar cost pressure, update-mode selection, and local metric shaping. Fixed-penalty PPO remains a reward-shaping reference baseline rather than a MACF-X adapter because its coefficient is manually fixed outside the constraint-feedback loop.

\subsection{MACF-PPO: adaptive Lagrangian integration}

MACF-PPO uses the PPO clipped surrogate as the backbone, but replaces learned dual ascent with the MACF-X pressure in Eq.~\eqref{eq:macfx-lambda}. For each rollout, define the effective advantage
\begin{equation}
  A_t^{\mathrm{MACF\text{-}PPO}}
  =
  A_t^r
  -
  \sum_{k\in\Kset}
  \lambda_k^{\mathrm{MACF}} A_t^{c,k}.
  \label{eq:macfx-ppo-adv}
\end{equation}
The policy is then updated with the standard PPO clipped objective
\begin{equation}
  \max_{\theta}
  \E_t
  \left[
    \min
    \left(
      q_t(\theta)A_t^{\mathrm{MACF\text{-}PPO}},
      \mathrm{clip}(q_t(\theta),1-\epsilon_{\mathrm{clip}},1+\epsilon_{\mathrm{clip}})
      A_t^{\mathrm{MACF\text{-}PPO}}
    \right)
  \right],
  \label{eq:macfx-ppo-objective}
\end{equation}
where $q_t(\theta)=\pi_\theta(a_t|o_t)/\pi_{\theta_{\mathrm{old}}}(a_t|o_t)$. The critic predicts one reward value and $|\Kset|$ cost values, and is trained by mean-squared error against reward and cost returns.

MACF-PPO is included because PPO-style updates are widely used in financial RL and provide a strong reward-optimization substrate. The MACF-X modification is minimal: the architecture and PPO update remain unchanged, while the static or dual-ascent penalty is replaced by a slack-aware, uncertainty-aware MACF pressure.

\begin{algorithm}[!htbp]
\caption{MACF-PPO}
\label{alg:macfx-ppo}
\begin{algorithmic}[1]
\State Collect rollout under $\pi_{\theta_{\mathrm{old}}}$ using financial observations $o_t$
\State Compute MACF costs $\mathbf c_t$, uncertainties $\bar{\mathbf u}_t$, reward advantages $A_t^r$, and cost advantages $A_t^{c,k}$
\State Estimate $\hat c_k$ and $\hat u_k$ using Eq.~\eqref{eq:appendix-vhat}
\State Compute $\lambda_k^{\mathrm{MACF}}$ using Eq.~\eqref{eq:macfx-lambda}
\State Form $A_t^{\mathrm{MACF\text{-}PPO}} = A_t^r-\sum_k\lambda_k^{\mathrm{MACF}}A_t^{c,k}$
\For{PPO epoch and minibatch}
  \State Update $\theta$ with the clipped PPO objective in Eq.~\eqref{eq:macfx-ppo-objective}
\EndFor
\State Update reward and cost critics by MSE
\end{algorithmic}
\end{algorithm}
\FloatBarrier

\subsection{MACF-CRPO: alternating-update integration}

Standard CRPO alternates between reward improvement when all constraints are feasible and cost reduction when a constraint is violated. MACF-CRPO preserves this alternating identity but changes both the switching rule and the constraint direction. Instead of waiting only for $\hat c_k>b_k$, it enters constraint mode when either a hard violation occurs or the MACF-X pressure becomes large:
\begin{equation}
  \mathrm{constraint\ mode}
  =
  \mathbf 1
  \left[
    \max_k(\hat c_k-b_k)>0
    \ \mathrm{or}\
    \max_k \lambda_k^{\mathrm{MACF}} \ge \lambda_{\mathrm{thr}}
  \right].
  \label{eq:macfx-crpo-switch}
\end{equation}
If the update is in reward mode, MACF-CRPO uses $A_t^r$. If the update is in constraint mode, it uses a normalized multi-head cost direction
\begin{equation}
  A_t^{\mathrm{MACF\text{-}CRPO}}
  =
  -
  \sum_{k\in\Kset}
  \bar\lambda_k A_t^{c,k},
  \qquad
  \bar\lambda_k
  =
  \frac{\lambda_k^{\mathrm{MACF}}}{\sum_{\ell\in\Kset}\lambda_\ell^{\mathrm{MACF}}+\epsilon_{\mathrm{safe}}}.
  \label{eq:macfx-crpo-adv}
\end{equation}
The selected advantage is then optimized with the same clipped PPO update used by the CRPO baseline.

MACF-CRPO is included to test whether the MACF field can improve first-order feasibility switching methods without introducing trust-region second-order machinery. It uses the same MACF-X pressure as MACF-PPO, but interprets it as an anticipatory switching signal and as a multi-head cost-reduction direction.

\begin{algorithm}[!htbp]
\caption{MACF-CRPO}
\label{alg:macfx-crpo}
\begin{algorithmic}[1]
\State Collect rollout and compute $A_t^r$, $A_t^{c,k}$, $\hat c_k$, and $\hat u_k$
\State Compute $\lambda_k^{\mathrm{MACF}}$ using Eq.~\eqref{eq:macfx-lambda}
\If{$\max_k(\hat c_k-b_k)\le 0$ and $\max_k\lambda_k^{\mathrm{MACF}}<\lambda_{\mathrm{thr}}$}
  \State Set update advantage $A_t \gets A_t^r$
  \State Set mode to reward improvement
\Else
  \State Set $\bar\lambda_k\gets \lambda_k^{\mathrm{MACF}}/(\sum_\ell\lambda_\ell^{\mathrm{MACF}}+\epsilon_{\mathrm{safe}})$
  \State Set update advantage $A_t \gets -\sum_k\bar\lambda_k A_t^{c,k}$
  \State Set mode to constraint reduction
\EndIf
\For{PPO epoch and minibatch}
  \State Update $\theta$ with the clipped PPO objective using $A_t$
\EndFor
\State Update reward and cost critics by MSE
\end{algorithmic}
\end{algorithm}
\FloatBarrier

\subsection{MACF-TRPO: safe geometry trust-region integration}

MACF-TRPO uses the trust-region interface directly. Let
\begin{equation}
  g_r
  =
  \nabla_\theta
  \E_t[q_t(\theta)A_t^r]_{\theta=\theta_{\mathrm{old}}},
  \qquad
  j_k
  =
  \nabla_\theta
  \E_t[q_t(\theta)A_t^{c,k}]_{\theta=\theta_{\mathrm{old}}},
  \label{eq:reward-cost-grads}
\end{equation}
and let $F$ denote the Fisher matrix induced by the rollout policy. MACF-TRPO defines the local safe metric
\begin{equation}
  G_{\mathrm{safe}}
  =
  F
  +
  \xi I
  +
  \sum_{k\in\Kset}
  \lambda_k^{\mathrm{MACF}} j_k j_k^\top ,
  \label{eq:safe-metric}
\end{equation}
where $\xi>0$ is a damping coefficient for numerical stability.
The normal-mode step is the reward-improving direction under this metric:
\begin{equation}
  d_r
  =
  G_{\mathrm{safe}}^{-1}g_r,
  \qquad
  \Delta\theta
  =
  \sqrt{
    \frac{2\delta}{d_r^\top G_{\mathrm{safe}}d_r}
  }d_r.
  \label{eq:macfx-trpo-step}
\end{equation}
The implementation computes products with $G_{\mathrm{safe}}$ through Fisher-vector products and the low-rank term in Eq.~\eqref{eq:safe-metric}, then solves Eq.~\eqref{eq:macfx-trpo-step} by conjugate gradient. A backtracking line search accepts a step only if the KL trust-region condition is satisfied and the linearized constraint checks
\begin{equation}
  \hat c_k + j_k^\top\Delta\theta \le b_k,
  \qquad k\in\Kset,
  \label{eq:linearized-constraint-check}
\end{equation}
hold.

If any head violates its budget, MACF-TRPO enters recovery mode. Recovery minimizes the violated cost heads under a smaller Fisher trust region. With deficits
\begin{equation}
  \omega_k
  =
  \frac{\max(0,\hat c_k-b_k)}
  {\sum_\ell \max(0,\hat c_\ell-b_\ell)+\epsilon_{\mathrm{safe}}},
  \label{eq:recovery-deficit-weights}
\end{equation}
the recovery direction is
\begin{equation}
  d_c
  =
  F^{-1}
  \sum_{k:\hat c_k>b_k}\omega_k j_k,
  \qquad
  \Delta\theta_{\mathrm{rec}}
  =
  -
  \sqrt{
    \frac{2\delta_{\mathrm{rec}}}{d_c^\top Fd_c}
  }d_c.
  \label{eq:macfx-trpo-recovery}
\end{equation}
Recovery exits only after all heads have slack relative to their budgets.

MACF-TRPO is included because trust-region methods provide the cleanest geometric interface for learned constraints: MACF does not change the reward, but changes the local geometry in which reward-improving steps are measured.

\begin{algorithm}[!htbp]
\caption{MACF-TRPO}
\label{alg:macfx-trpo}
\begin{algorithmic}[1]
\State Collect rollout and compute $A_t^r$, $A_t^{c,k}$, $\hat c_k$, and $\hat u_k$
\State Compute $g_r$, cost gradients $j_k$, Fisher-vector product oracle $Fv$, and $\lambda_k^{\mathrm{MACF}}$
\If{all $\hat c_k\le b_k$ and not in recovery}
  \State Define $G_{\mathrm{safe}}v = Fv+\xi v+\sum_k\lambda_k^{\mathrm{MACF}}j_k(j_k^\top v)$
  \State Solve $G_{\mathrm{safe}}d_r=g_r$ by conjugate gradient
  \State Scale $d_r$ to KL radius $\delta$
  \State Backtrack until reward improves, KL is valid, and Eq.~\eqref{eq:linearized-constraint-check} holds
\Else
  \State Compute deficit weights $\omega_k$ over violated heads
  \State Solve $Fd_c=\sum_{k:\hat c_k>b_k}\omega_k j_k$
  \State Backtrack along $-d_c$ until KL is valid and the weighted violated-cost surrogate decreases
  \State Exit recovery only when all heads have slack
\EndIf
\State Update reward and cost critics by MSE
\end{algorithmic}
\end{algorithm}
\FloatBarrier

\subsection{MACF-CPO: safe geometry recovery integration}

MACF-CPO uses the same normal-mode update as MACF-TRPO, but makes the recovery step closer to CPO-style feasibility correction. In recovery, MACF-CPO aggregates violated heads using both deficit and MACF-X pressure:
\begin{equation}
  \omega_k^{\mathrm{CPO}}
  =
  \frac{
    \max(0,\hat c_k-b_k)\lambda_k^{\mathrm{MACF}}
  }{
    \sum_{\ell:\hat c_\ell>b_\ell}
    \max(0,\hat c_\ell-b_\ell)\lambda_\ell^{\mathrm{MACF}}
    +\epsilon_{\mathrm{safe}}
  }.
  \label{eq:macfx-cpo-omega}
\end{equation}
Unlike MACF-TRPO recovery, which uses the Fisher geometry, MACF-CPO uses the same safe metric in Eq.~\eqref{eq:safe-metric} for feasibility correction:
\begin{equation}
  d_c^{\mathrm{CPO}}
  =
  G_{\mathrm{safe}}^{-1}
  \sum_{k:\hat c_k>b_k}\omega_k^{\mathrm{CPO}}j_k,
  \qquad
  \Delta\theta_{\mathrm{rec}}^{\mathrm{CPO}}
  =
  -
  \sqrt{
    \frac{2\delta_{\mathrm{rec}}}
    {(d_c^{\mathrm{CPO}})^\top G_{\mathrm{safe}}d_c^{\mathrm{CPO}}}
  }d_c^{\mathrm{CPO}}.
  \label{eq:macfx-cpo-recovery}
\end{equation}
This makes both reward-improvement and feasibility-recovery steps aware of the learned MACF cost geometry.

MACF-CPO is included to represent CPO-style methods, where feasibility correction is a first-class part of the algorithm rather than an external penalty. It tests whether the same MACF-X geometry remains useful when the optimizer explicitly alternates between reward improvement and feasibility recovery.

\begin{algorithm}[!htbp]
\caption{MACF-CPO}
\label{alg:macfx-cpo}
\begin{algorithmic}[1]
\State Collect rollout and compute $A_t^r$, $A_t^{c,k}$, $\hat c_k$, and $\hat u_k$
\State Compute $g_r$, $j_k$, $\lambda_k^{\mathrm{MACF}}$, and $G_{\mathrm{safe}}$
\If{all $\hat c_k\le b_k$ and not in recovery}
  \State Take the MACF-TRPO normal step under $G_{\mathrm{safe}}$
\Else
  \State Compute $\omega_k^{\mathrm{CPO}}$ using Eq.~\eqref{eq:macfx-cpo-omega}
  \State Solve $G_{\mathrm{safe}}d_c^{\mathrm{CPO}}=\sum_{k:\hat c_k>b_k}\omega_k^{\mathrm{CPO}}j_k$
  \State Step along $-d_c^{\mathrm{CPO}}$ under recovery radius $\delta_{\mathrm{rec}}$
  \State Backtrack until KL is valid and the weighted violated-cost surrogate decreases
\EndIf
\State Update reward and cost critics by MSE
\end{algorithmic}
\end{algorithm}
\FloatBarrier

\subsection{Framework consistency}

All MACF-X variants share the same three design commitments. First, the portfolio policy remains financial-only: ESG evidence enters only through MACF costs and not through the policy observation or reward. Second, every optimizer receives the same head-wise constraint object $(A_t^{c,k},\hat c_k,\hat u_k,b_k)$, derived from the add-risk, hold-risk, and spillover-risk heads in the main text. Third, the MACF-X pressure in Eq.~\eqref{eq:macfx-lambda} is the single translation layer from learned ESG evidence to optimization behavior. In PPO it becomes adaptive cost pressure; in CRPO it becomes an anticipatory switching and weighting signal; in TRPO and CPO it becomes a local geometry. This preserves a common conceptual framework while allowing each constrained-RL interface to use the learned constraint field through its natural update interface.

\section{Universe construction and cross-market matching}
\label{app:universe-construction}

This appendix documents how the US30 and EU30 universes are constructed. The goal of the experimental universe is not to approximate a full investable benchmark mechanically, but to create a controlled setting in which ESG evidence is sufficiently observable, sector exposure is balanced, and cross-market comparisons are interpretable.

\paragraph{US30 construction.}
The US30 universe contains 30 large U.S. equities, organized as six sectors with five firms per sector. We selected sectors to cover heterogeneous ESG sensitivity rather than only the highest-emission industries. Energy, Utilities, and Industrials provide direct exposure to environmental, labor, safety, and transition-risk channels. Financials provide governance, stewardship, and financing-channel exposure. Consumer Staples and Information Technology provide comparatively less mechanically emission-driven, but highly visible, disclosure-rich settings where social, supply-chain, governance, data, and labor controversies can be observed. Within each sector, we use large, liquid firms because they have denser news coverage, richer public disclosure, more frequent analyst and institutional attention, and are more likely to appear in portfolio-level ESG mandates. This design is better suited to evaluating point-in-time ESG constraints than a broad index sample in which many firms may have sparse evidence over the test window.

The balanced sector design also avoids a common confound in ESG-aware portfolio experiments. If the universe is taken directly from a broad benchmark such as the S\&P 500, sector composition, market capitalization, disclosure intensity, and news availability vary substantially across industries. A method may then appear to improve ESG behavior simply by exploiting sector imbalance or evidence-density imbalance. The US30 design fixes the number of firms per sector, making comparisons across optimizers less dependent on index composition and more focused on how each method consumes the same learned constraint object.

\paragraph{EU30 matching protocol.}
The EU30 universe is built as a cross-market counterpart rather than as an independent benchmark. We keep the same six sector groups and select five large European firms in each group. The matching criterion is sector-level functional comparability, not one-to-one firm equivalence. For example, the European Energy group contains integrated oil and gas firms that provide a comparable ESG-risk context to the U.S. Energy group, while the European Utilities group contains major electricity and power-transition firms. This preserves the experimental structure of the US30 study while introducing a different market, disclosure regime, listing geography, and news environment. The EU30 study therefore tests whether MACF-X remains useful when the learned constraint pipeline is deployed on a matched but non-U.S. universe.

\begin{table*}[!htbp]
  \centering
  \scriptsize
  \caption{US30 and EU30 universe construction. Each universe contains six sectors with five large, liquid firms per sector. EU30 is matched at the sector and economic-function level rather than by one-to-one firm pairs.}
  \label{tab:universe-construction}
  \begin{tabular}{lll}
    \toprule
    Sector & US30 tickers & EU30 tickers \\
    \midrule
    Energy &
    XOM, CVX, COP, SLB, EOG &
    SHELL.AS, TTE.PA, ENI.MI, REP.MC, OMV.VI \\
    Industrials &
    CAT, DE, GE, UNP, RTX &
    SIE.DE, SU.PA, AIR.PA, SAF.PA, ALO.PA \\
    Consumer Staples &
    WMT, COST, PG, KO, PEP &
    UNA.AS, HEIA.AS, OR.PA, RI.PA, BN.PA \\
    Financials &
    JPM, BAC, GS, MS, BLK &
    SAN.MC, BNP.PA, ALV.DE, INGA.AS, UCG.MI \\
    Information Technology &
    AAPL, MSFT, NVDA, AVGO, QCOM &
    SAP.DE, ASML.AS, IFX.DE, CAP.PA, NOKIA.HE \\
    Utilities &
    NEE, SO, DUK, AEP, XEL &
    ENEL.MI, IBE.MC, RWE.DE, EOAN.DE, ENGI.PA \\
    \bottomrule
  \end{tabular}%
\end{table*}
\FloatBarrier

\paragraph{Data-coverage rationale.}
The universe construction is tied to the MACF learning problem. MACF requires point-in-time multimodal evidence, including firm-level events, peer pressure, anchor-confirmed incidents, and portfolio context. Large firms are therefore not chosen only for liquidity; they are chosen because their ESG-relevant actions are more likely to leave observable public traces. In the processed data used by the experiments, the US30 universe has complete ticker coverage in both news and anchor evidence, while the EU30 robustness branch is constructed to mirror the same 30-firm, six-sector schema. This coverage criterion is important because the paper evaluates learned constraints, not merely static ESG scores. Sparse evidence would make the experiment primarily a missing-data study rather than a test of action-conditioned ESG constraint integration.

The SP500-scale branch is useful as an engineering target, but it is not used as the main experimental universe in this version because broad-index coverage is uneven across tickers and sectors. Using it as the headline setting would make evidence availability, sector weights, and market-cap composition difficult to disentangle from the optimizer effects. The US30/EU30 design instead emphasizes controlled sector balance, high-disclosure firms, and matched cross-market structure, which is more appropriate for evaluating whether MACF-X can inject learned ESG constraints into existing portfolio DRL optimizers without changing the financial reward.

\paragraph{External data sources and terms.}
Market price and volume data used in this study are sourced from public financial-data providers under their applicable terms of use. ESG evidence is collected from public regulatory filings, official-source material, and major news sources; the research-use data construction pipeline preserves source attribution and access metadata so that derived evidence records can be audited against their original public sources.

\section{Financial reward and portfolio-control protocol}
\label{app:financial-protocol}

This appendix details the financial reward and action constraints used in the experiments. The main purpose is to make clear that MACF-X is evaluated under a common portfolio-control protocol: all methods optimize the same financial reward, observe the same financial state, face the same trading frictions, and differ only in how they consume the learned ESG constraint signal.

\paragraph{ESG-free financial reward.}
The reward is deliberately financial-only. For a one-period portfolio return $r_t^{\mathrm{port}}$, weight change $\Delta w_t$, and drawdown $dd_t$, the implemented reward is
\begin{equation}
  r_t
  =
  \log(1+r_t^{\mathrm{port}})
  -
  c_{\mathrm{tx}}\sum_{i=1}^{N}|\Delta w_{i,t}|
  -
  \lambda_{\mathrm{dd}}\max(0, |dd_t|-d_0).
  \label{eq:appendix-financial-reward}
\end{equation}
The first term uses log portfolio return, the second term charges proportional transaction cost on turnover, and the third term penalizes drawdowns only after a tolerance threshold. ESG costs do not enter Eq.~\eqref{eq:appendix-financial-reward}. They are returned separately as the MACF constraint vector and are consumed only by constrained optimizers. This separation is important because it prevents the experiments from rewarding ESG exposure directly or treating ESG as an alpha factor.

\paragraph{Trading frictions and risk-control terms.}
The main US30 and EU30 MACF-X experiments use $c_{\mathrm{tx}}=0.0005$, corresponding to five basis points per unit turnover. The drawdown penalty uses $\lambda_{\mathrm{dd}}=0.02$ and $d_0=0.05$, so small fluctuations are not penalized, while larger drawdowns reduce the reward. These parameters are shared across methods within each reported experiment. Return, Sharpe, Calmar, turnover, and ESG Violation are computed from the same deterministic test rollouts, so no method receives a more favorable financial evaluation rule.

\paragraph{Cash and position constraints.}
The action is represented as target portfolio weights over $N$ risky assets plus cash. The raw policy output is converted into an invested fraction $\alpha_t$ and a normalized risky-asset allocation. In the reported experiments,
\begin{equation}
  \alpha_t \in [0.85,1],
\end{equation}
so the portfolio must keep at least $85\%$ invested and can hold at most $15\%$ in cash. This prevents constrained methods from satisfying ESG budgets by simply exiting the equity market. The same action converter also applies a single-name cap of $15\%$ and a sector cap of $35\%$. These caps are financial portfolio-construction constraints, not ESG constraints, and they are applied equally to unconstrained, baseline constrained, and MACF-X methods.

\paragraph{Turnover deadband and target-weight execution.}
To avoid unrealistic micro-trading, the environment applies a no-trade deadband: target-weight changes smaller than $\epsilon_{\mathrm{trade}}=0.002$ are treated as no trade for that asset. After the action is converted to feasible target weights, the environment computes turnover as $\sum_i |\Delta w_{i,t}|$, charges transaction cost through Eq.~\eqref{eq:appendix-financial-reward}, and then evaluates the next-period portfolio return from the same market data for all methods. MACF receives the signed post-conversion weight change, ensuring that the learned ESG cost is tied to the actual contemplated portfolio transition rather than to an unconstrained raw action.

\paragraph{Fair-comparison principle.}
The protocol is designed so that differences in reported performance cannot be attributed to different market observations, different financial rewards, different cash permissions, or different transaction-cost assumptions. All methods use the same financial-only observation space, the same train/validation/test split, the same minimum-invested rule, the same position caps, the same turnover cost, and the same drawdown penalty. Fixed-penalty PPO is the only baseline that inserts ESG cost directly into the reward-side objective, and it is therefore reported as a reward-shaping reference rather than as a MACF-X adapter. The MACF-X variants leave the financial reward unchanged and inject ESG only through the constraint interface of the corresponding optimizer.

\section{MACF supervision and training protocol}
\label{app:macf-training}

This appendix summarizes how the MACF constraint field is learned. MACF is trained before policy optimization with weak supervision constructed from point-in-time ESG evidence, then frozen and used by all portfolio optimizers. Thus, policy-gradient updates never train MACF and ESG information never enters the financial reward.

\paragraph{Training samples.}
For each date $t$, ticker $i$, and candidate weight change, we construct one MACF input $H_{i,t}$ from the pipeline in Appendix~\ref{app:data-pipeline}. The signed post-conversion target-weight change allows the same firm evidence to imply different costs for increasing, maintaining, or reducing exposure.

\paragraph{Weak-supervision targets.}
MACF is not trained to reproduce a scalar ESG score. Each sample receives three continuous risk targets
\[
  (y^{\mathrm{add}},y^{\mathrm{hold}},y^{\mathrm{spill}})\in[0,1]^3
\]
and three uncertainty targets
\[
  (u^{\mathrm{add}},u^{\mathrm{hold}},u^{\mathrm{spill}})\in[0,1]^3.
\]
These targets are deterministic weak-labeling functions of point-in-time evidence rather than ground-truth ESG costs. Add-risk combines recent firm-specific severity, recency, anchor confirmation, repeated high-severity events, and the signed weight change, so exposure increases under active controversies receive higher labels than reductions. Hold-risk summarizes persistent ownership exposure through unresolved events, event frequency, mean severity, and time since severe evidence. Spillover-risk uses peer pressure, recent peer high-severity activity, and stressed macro regimes to capture sector or market transmission. Uncertainty targets encode evidence coverage and confirmation: sparse or weakly anchored evidence receives higher uncertainty, while official or strongly anchored evidence lowers uncertainty. The numerical weights used inside these labeling functions are fixed coefficients; they are used only to construct weak labels and are held fixed before policy optimization and test evaluation. The complete labeling implementation is provided in the anonymized repository.

\paragraph{Model and loss.}
MACF uses z-scored numerical features, one-hot categorical features, event/text features, a shared MLP backbone, and three head-specific output layers. Each head outputs
\[
  (\rho^{\Delta,k},\rho^{w,k},u^k)\in[0,1]^3.
\]
The exposure operators are trained sparsely: the add-risk head trains $\rho^{\Delta,\mathrm{add}}$ toward $y^{\mathrm{add}}$ and its complementary hold-output toward zero, while the hold-risk and spillover-risk heads train $\rho^{w,k}$ toward $y^k$ and their complementary add-output toward zero. For a minibatch $\mathcal B$, the supervised loss is
\begin{equation}
\begin{aligned}
  \mathcal L_{\mathrm{sup}}
  &=
  \frac{1}{|\mathcal B|}
  \sum_{(i,t,\Delta w_{i,t})\in\mathcal B}
  \sum_{k\in\Kset}
  \Bigl[
    \|\rho_{\mathrm{prim}}^{k}-y^{k}\|_2^2
    +0.1\|\rho_{\mathrm{res}}^{k}\|_2^2
    +0.4\|u^k-\tilde u^k\|_2^2
  \Bigr],
\end{aligned}
  \label{eq:macf-supervised-loss}
\end{equation}
where $\rho_{\mathrm{prim}}^{k}$ is the exposure output associated with the target mechanism, $\rho_{\mathrm{res}}^{k}$ is the complementary residual exposure output, and $\tilde u^k$ is the uncertainty target.

We add a soft monotonicity regularizer for add-risk. For a fixed $(i,t)$, order signed weight changes from exposure reduction to exposure increase and penalize decreases in add-risk as the transition becomes more purchase-oriented:
\begin{equation}
  \mathcal L_{\mathrm{mono}}
  =
  \frac{1}{|\mathcal G|}
  \sum_{(i,t)\in\mathcal G}
  \sum_{j}
  \max
  \left(
    0,
    \rho_{i,t,j}^{\Delta,\mathrm{add}}
    -
    \rho_{i,t,j+1}^{\Delta,\mathrm{add}}
  \right).
  \label{eq:macf-mono-loss}
\end{equation}
The final MACF objective is
\begin{equation}
  \mathcal L_{\mathrm{MACF}}
  =
  \mathcal L_{\mathrm{sup}}
  +
  \lambda_{\mathrm{mono}}\mathcal L_{\mathrm{mono}},
  \qquad
  \lambda_{\mathrm{mono}}=0.2.
  \label{eq:macf-final-loss}
\end{equation}

\paragraph{Training protocol.}
MACF is trained with Adam (learning rate $10^{-3}$, weight decay $10^{-5}$, batch size $512$, at most $15$ epochs, early stopping patience $4$). Model selection uses validation loss on 2023 data, after which the selected MACF is frozen before policy optimization. The fitted encoder and selected checkpoint are used to generate constraint predictions for train, validation, and test rollouts. The scalar ESG risk score, when available, is used only as a diagnostic reconstruction check and not as a training target. Full hyperparameter configurations are also provided in the anonymized repository.


\end{document}